\documentclass[10pt,twocolumn,letterpaper]{article}

\usepackage{cvpr}
\usepackage{times}
\usepackage{epsfig}
\usepackage{graphicx}
\usepackage{amsmath}
\usepackage{amssymb}
\usepackage{multirow}

\usepackage[pagebackref=true,breaklinks=true,letterpaper=true,colorlinks,bookmarks=false]{hyperref}
\cvprfinalcopy 

\ifcvprfinal\fi
\begin{document}

\title{HetConv: Heterogeneous Kernel-Based Convolutions for Deep CNNs}


\author{
Pravendra Singh \hspace{1.2cm}Vinay Kumar Verma \hspace{1.2cm}Piyush Rai\hspace{1.2cm}Vinay P. Namboodiri\\
Department of Computer Science and Engineering, IIT Kanpur, India\\
{\tt\small \{psingh, vkverma, piyush, vinaypn\}@cse.iitk.ac.in}
}

\maketitle
\thispagestyle{empty}

\begin{abstract}
We present a novel deep learning architecture in which the convolution operation leverages heterogeneous kernels. The proposed HetConv (Heterogeneous Kernel-Based Convolution) reduces the computation (FLOPs) and the number of parameters as compared to standard convolution operation while still maintaining representational efficiency. To show the effectiveness of our proposed convolution, we present extensive experimental results on the standard convolutional neural network (CNN) architectures such as VGG \cite{vgg2014very} and ResNet \cite{resnet}. We find that after replacing the standard convolutional filters in these architectures with our proposed HetConv filters, we achieve 3X to 8X FLOPs based improvement in speed while still maintaining (and sometimes improving) the accuracy. We also compare our proposed convolutions with group/depth wise convolutions and show that it achieves more FLOPs reduction with significantly higher accuracy. 
\end{abstract}

\section{Introduction}
Convolutional neural networks \cite{resnet,krizhevsky2012imagenet,vgg2014very} have shown remarkable performance in domains like Vision and NLP. The general trend to improve performance further has made models more complex and deeper. Increasing the accuracy by increasing model complexity with a deeper network is not for free; it comes with the cost of a tremendous increase in computation (FLOPs). Therefore, various types of convolution operations/convolutional filters have been proposed to reduce FLOPs to the model more efficient. 

Existing convolutional filters can be roughly divided into three categories: 1- Depthwise Convolutional Filter to perform depthwise convolution (DWC) \cite{depthwise_iclr}, 2- Pointwise Convolutional Filter to perform pointwise convolution (PWC) \cite{szegedy2015googlenet} and 3- Groupwise Convolutional Filter to perform groupwise convolution (GWC) \cite{krizhevsky2012imagenet}. Most of the recent architectures \cite{howard2017mobilenets,szegedy2017inception,chollet2017xception,iandola2016squeezenet,szegedy2016rethinkinginception,zhangshufflenet} use a combination of these convolutional filters to make the model efficient. Using these convolutions (e.g., DWC, PWC, and GWC), many of the popular models \cite{iandola2016squeezenet, howard2017mobilenets,chollet2017xception} have explored new architectures to reduce FLOPs. However, designing a new architecture requires a lot of work to find out the best combination of filters that result in minimal FLOPs.

Another popular approach to increase the efficiency of a model is to use model compression \cite{binarycompression,chen2015compressing,thiNet17,channelPruning17,weightSumICLR-17,he2018soft,yu2017nisp}. Model compression can be broadly categorized into three categories: connection pruning \cite{han2015deep}, filter pruning \cite{thiNet17,channelPruning17,weightSumICLR-17,he2018soft,singh2018multi,singh2018stability,singh2018leveraging} and quantization \cite{han2015deep,binarycompression}. 

In filter pruning, the idea is to prune a filter that has the minimal contribution in the model, and after removing this filter/connection, the model is usually finetuned to maintain its performance. While pruning the model, we require a pre-trained model (possibly requiring a computationally expensive training as a preprocessing step), and then later we discard the filter that has a minimal contribution. Hence it is a very costly and tricky process. Therefore, using an efficient convolutional filter or convolution operation to design an efficient architecture is a  more popular approach than pruning. This does not require expensive training and then pruning since training is done from scratch efficiently.  

Using efficient convolutional filters, there are two different objectives. One kind of work focuses on designing architectures that have minimal FLOPs while compromising on accuracy. These works focus on developing the model for the IoT/low-end device \cite{howard2017mobilenets,zhangshufflenet}. These models suffer from the low accuracy hence they have to search the best possible model to create a balance between accuracy and FLOPs. So there is a tradeoff between FLOPs and model accuracy. 

Another set of work focuses on increasing accuracy while keeping the model FLOPs the same as the original architecture. The recent architectures, such as Inception \cite{szegedy2017inception}, RexNetXt \cite{resnetxt} and Xception \cite{chollet2017xception} are examples of this kind of work. Their objective is to design a more complex model using efficient convolutional filters while keeping the FLOPs the same as the base model. It is usually expected that a more complex model would learn better features, resulting in better accuracies. However, these methods are not focused on designing a new architecture, but primarily on using existing efficient filters in standard base architectures. Therefore these works keep the number of layers and the architecture the same as the base model and increase the filters on each layer such that it does not increase the FLOPs.

In contrast to these two approaches, the primary focus of our work is to reduce the FLOPs of the given model/architecture by designing new kernels, without compromising on the loss of accuracy.
Experimentally we find that the proposed approach has much lower FLOPs than the state-of-art pruning approaches while maintaining the accuracy of the base model/architecture. The pruning approaches are very costly and show a significant drop in accuracy to achieve FLOPs compression.

In the proposed approach, we are choosing a different strategy to increase the efficiency of the existing model without sacrificing the accuracy. An architecture search requires years of research to get an optimized architecture. Therefore, instead of designing a new efficient architecture, we design an efficient convolution operation (convolutional filter) that can be directly plugged into any existing standard architecture to reduce FLOPs. To achieve this, we propose a new type of convolution - \emph{heterogeneous} convolution.  


The convolution operation can be divided into two categories based on the types of the kernel:
\begin{itemize}
  \item Homogeneous convolution using a traditional convolutional filter (for example standard convolution, groupwise convolution, depthwise convolution, pointwise convolution). Homogeneous convolution can be performed using a homogeneous filter. A filter is said to be homogeneous if it contains all kernels of the same size (for example, in a $3\times3\times256$ CONV2D filter, all 256 kernels will be of size $3\times3$). 
  \item Heterogeneous convolution uses a heterogeneous convolutional filter (HetConv). A filter is said to be heterogeneous if it contains different sizes of kernels (for example, in a HetConv filter, out of 256 kernels some kernels are of size $3\times3$ and remaining kernels are of size $1\times1$).
\end{itemize}

Using a heterogeneous filter in deep CNN overcomes the limitation of the existing approaches that are based on efficient architecture search and model compression. One of the latest efficient architecture MobileNet \cite{howard2017mobilenets} uses depthwise and pointwise convolution. The standard convolutional layer is replaced by two convolutional layers hence it has more latency (latency one). Please refer to Section-3.3 and Figure-\ref{fig:latency} for more details about latency. But our proposed HetConv has same latency as the original architecture (latency zero) unlike \cite{howard2017mobilenets,szegedy2017inception,szegedy2015googlenet,chollet2017xception} that have latency greater than zero. 

Compared to model compression that suffers from high accuracy drop, our approach is very competitive to the state-of-art result of the standard model like ResNet \cite{resnet} and VGGNet \cite{vgg2014very}. Using HetConv filters, we can train our model from scratch, unlike pruning approaches that need a pre-trained model, without sacrificing accuracy. The pruning approaches also suffer from sharp accuracy drop if we increase the degree of FLOP pruning. Using proposed HetConv filters, we have state-of-art result regarding FLOPs compare to the FLOP pruning methods. Also, the pruning process is inefficient as it takes a lot of time in training and fine tuning after pruning. Our approach is highly efficient and gives a similar result compared to the original model while training from scratch.  

To the best of our knowledge, this is the first convolution/filter that is heterogeneous. This heterogeneous design helps to increase the efficiency (FLOPs reduction) of the existing architecture without sacrificing the accuracy. We did extensive experiment on different architectures like ResNet \cite{resnet}, VGG-16 \cite{vgg2014very} etc just by replacing their original filters to our proposed filters. We found that without sacrificing the accuracy of these models, we have a high degree of FLOPs reduction (3X to 8X). These FLOPs reductions are even significantly better as compared to existing pruning approach. 

Our main contributions are as follows:
\begin{itemize}
    \item We design an efficient heterogeneous convolutional filter, that can be plugged into any existing architecture to increase the efficiency (FLOPs reduction of order 3X to 8X) of the architecture without sacrificing the accuracy. 
    \item The proposed HetConv filters are designed in such a way that it has zero latency.  Therefore, there is negligible delay from input to output. 
\end{itemize}

\section{Related Work}
The recent success of deep neural network \cite{krizhevsky2012imagenet,ren2015fasterrcnn,resnet,huang2017densely,finn2017model,goodfellow2014generative,noroozi2016jigsaw,verma2018generalized,snell2017prototypical} depends on the model design. To achieve a minimal error rate, the model becomes more and more complex. The complex and deeper architecture contain millions of parameters and requires billions of FLOPs (computations) \cite{resnet,vgg2014very,huang2017densely}. These models require machines with high-end specifications, and these type of architecture are very inefficient on low computing resources. This raises interest in designing efficient models \cite{he2015convolutional}. The work to increase the efficiency of the model can be divided into two parts.

\subsection{Efficient Convolutional Filter}
To design the efficient architecture recently few novel convolutional filters have been proposed. Among them Groupwise Convolution (GWC) \cite{krizhevsky2012imagenet},  Depthwise convolution (DWC) \cite{depthwise_iclr} and Pointwise Convolution (PWC) \cite{resnet} are the popular convolutional filters. These are widely used to design efficient architecture. GoogleNet \cite{szegedy2015googlenet} use the inception module and irregular stacking architecture. Inception module uses GWC and PWC to reduce FLOPs. ResNet \cite{resnet,he2016identity} uses a bottleneck structure to design an efficient architecture with  residual connection. They use  PWC and the standard convolution that help to go deeper without increasing the model parameter and reduces the FLOPs explosion. Therefore they can design a much deeper architecture compare to VGG \cite{vgg2014very}. ResNetxt \cite{resnetxt} use the ResNet architecture and they divide each layer with GWC and PWC.  Therefore without increasing FLOPs, they can increase the cardinality \footnote{The size of the set of transformations}. They show that increasing cardinality is much more effective than a deeper or wider network. SENet \cite{hu2017squeeze} design a new connection that gives the weight to each output feature map with a minor increase in FLOPs but shows a boost in the performance.

MobileNet \cite{howard2017mobilenets} is another popular architecture specially designed for the IoT devices contains DWC and PWC. This architecture is very light and highly efficient in term of FLOPs. This reduction in FLOPs are not for free and come with the cost of a drop in the accuracy compared to the state-of-art models. \cite{ioannou2015training,ioannou2017deep} use different types of convolutional filers at the same layers, but each filter performs a \emph{homogeneous} convolution due to the presence of same types of kernels in each filter. Using different types of convolutional filers at the same layers also helps in reducing parameters/FLOPs. In our proposed convolution, the convolution operation is heterogeneous due to the presence of \emph{different types of kernels in each filter}.  

\subsection{Model Compression}
Another popular approach to increase the efficiency of CNN is model compression. These can be categorised as: 1- Connection Pruning \cite{han2015deep,zhang2015efficient}, 2- Filter Pruning \cite{louizos2017bayesian,ding2018auto,singh2018multi,singh2018stability,singh2018leveraging} and 3- Bit Compression \cite{binarycompression}.  Filter pruning approaches are more effective as compared to other approaches and give high compression rate in terms of FLOPs. Also, the filter pruning approaches do not need any special hardware/software support (sparse library). 

Most of the works in filter pruning calculates the importance of the filter and prunes them based on some criteria followed by re-training to recover the accuracy drop.  \cite{li2016pruning} used $l_1$ norm as a metric for ranking filters. But the pruning is done on the pre-trained model and involves iterative training and the pruning which is costly. Also, filter pruning shows a sharp accuracy drop in accuracy, if the degree of flop pruning increases \cite{ding2018auto,yu2017nisp, thiNet17}.

\section{Proposed Method}
\begin{figure}[t]
    \centering
    \includegraphics[scale=0.53]{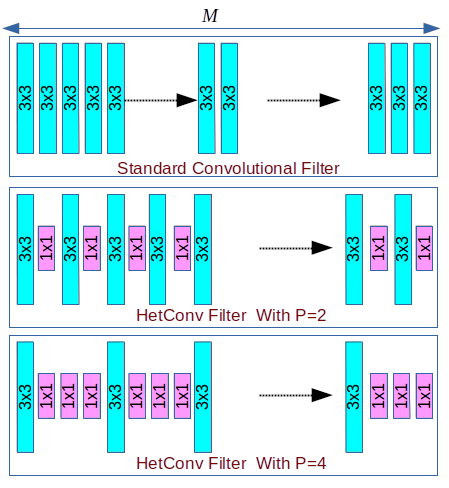}
    \caption{Difference between standard convolutional filter (homogeneous) and heterogeneous convolutional filter (HetConv). Here M is the input depth (number of input channels), and P is the part. Out of M kernels, M/P kernels will be of size $3\times 3$ and remaining will be $1\times 1$ kernels.}
    \label{fig:difference}
\end{figure}
 
\begin{figure*}[t]
    \centering
    \includegraphics[width=17.5cm,height=5cm]{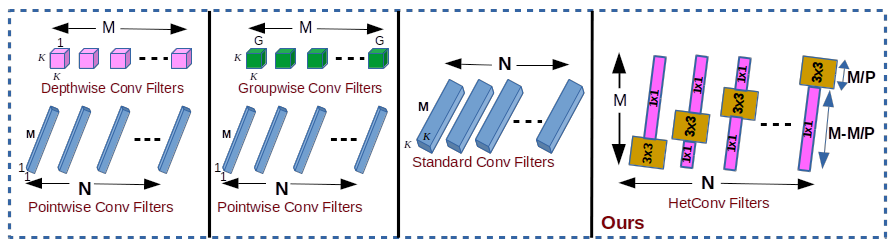}
    \caption{Comparison between the proposed Convolutional filters (HetConv) with other efficient convolutional filters. Our heterogeneous filters have zero latency while other (GWC+PWC or DWC+PWC) have a latency of one unit.}
    \label{fig:compare}
\end{figure*}

In this work, we propose a novel filter/convolution (HetConv) that contains a heterogeneous kernel (e.g., few kernels are of size $3\times3$, and others may be $1\times1$) to reduce the FLOPs of existing models with the same accuracy as the original model. This is very different from the standard convolutional filter that is made of homogeneous kernels (say all $3\times3$ or all $5\times5$). The heterogeneous filter is very efficient in terms of FLOPs. It can be approximated as a combined filter of a groupwise convolutional filter (GWC) and pointwise convolutional filter (PWC). To reduce the FLOPs of a convolutional layer, we generally replace it by two or more layers (GWC/DWC and PWC), but it increases the latency because next layer's input is the previous layer's output. Hence all computations have to be done sequentially to get the correct output. In contrast, our proposed HetConv has the same latency. Difference between the standard filter and HetConv filter is shown in the Figure- \ref{fig:difference} and Figure- \ref{fig:compare}.

In the standard convolutional layer, let us assume input (input feature map) of size $D_i \times D_i \times M$. Here $D_i$ is the input square feature map spatial width and height and $M$ is the input depth (number of input channels). Also consider $D_o \times D_o \times N$ is the output feature map. Here $D_o$ is the output square feature map spatial width and height and $N$ is the output depth (number of output channels). An output feature map is obtained by applying the $N$ filters of size $K \times K \times M$. Here K is the kernel size. Therefore the total computational cost at this layer L can be given as:
\begin{equation}\label{eq:standardconv}
    FL_S= D_o \times D_o \times M \times N \times K \times K
\end{equation}

 It is clear from the Equation-\ref{eq:standardconv} the computational cost depends on the kernel size (K), feature map size, input channels $M$ and output channels $N$. This computational cost is very high which can be further reduced by carefully designing the new convolution operation. To reduce the high computation, various convolutions like DWC, PWC and GWC are proposed which is used in the many recent architecture \cite{howard2017mobilenets,zhangshufflenet,szegedy2017inception} to reduce the FLOPs but all of them increase the latency.

\begin{figure}[htb!]
    \centering
    \includegraphics[width=8.2cm,height=5cm]{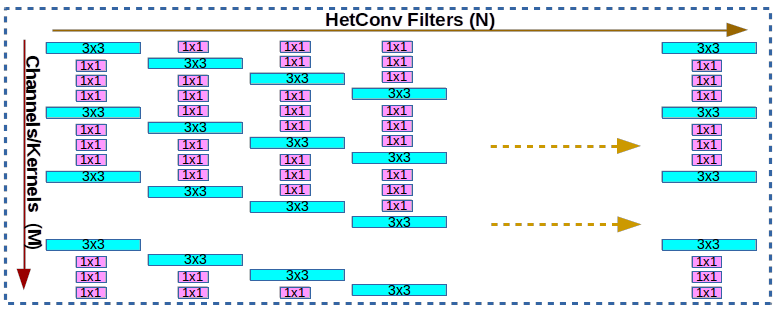}
    \caption{\textbf{Convolutional filters at a layer L:} Proposed Convolutional Filter (HetConv) using Heterogeneous Kernel. In this Figure, each channel is made of using the heterogeneous kernel of size $3\times3$ and $1\times1$. Replacing $3\times3$ kernels with $1\times1$ kernels in standard convolutional filter reduces the FLOPs dramatically while maintaining the accuracy. Filters of a particular layer are arranged in a shifted manner (i.e., if the first filter starts $3\times 3$ kernel from the first position then the second filter starts the $3\times 3$ kernel from the second position and so on).}
    \label{fig:hetro}
\end{figure}

 The standard convolution operation and some recent convolution operations \cite{howard2017mobilenets,zhangshufflenet,szegedy2017inception} use a homogeneous kernel (i.e., each kernel is of the same size for the whole filter). Here to increase the efficiency we are using the heterogeneous kernels. This contains different size kernels for the same filter. Please refer to Figure-\ref{fig:hetro} to visualize all filters at a particular layer L.  Let us define part $P$ which controls the number of different types of kernels in a convolutional filter. For part $P$, a fraction $1/P$ out of total kernels will be for $K \times K$ kernels and remaining fraction $(1-1/P)$ will be for $1\times1$ kernels. For better understanding, Let's take an example, in a $3\times3\times256$ standard CONV2D filter if you replace $(1-1/P)*256$, $3\times3$ kernels with $1\times1$ (along with the central axis), you will get a HetConv filter with part $P$. Please refer to  Figure-\ref{fig:difference} and  \ref{fig:compare}. 
 
 The computational cost for $K\times K$ size kernels in the HetConv filters with part $P$ on the layer L  is given as:
 
 \begin{equation}
     FL_K = (D_o \times D_o \times M \times N \times K \times K)/P
 \end{equation}
It reduces the cost $P$  times since instead of $M$ kernels of size $K\times K$, now we have only $M/P$ kernels of size $K\times K$. 
 
The remaining $(M-M/P)$ kernels are of size $1\times 1$. The computational cost of the remaining $1\times1$ kernels can be given as:
\begin{equation}
     FL_1= (D_o \times D_o  \times N )\times \left( M-\frac{M}{P}\right)
\end{equation}
Therefore the total computational cost at layer L is given as:
\begin{equation}
    FL_{HC} = FL_K + FL_1
\end{equation}

The total reduction (R) in the computation as compared to standard convolution can be given as:

\begin{equation}\label{eq:speedup}
    \begin{split}
    R_{HetConv}=&\frac{FL_K+FL_1}{FL_S}\\
    =&\frac{1}{P}+\frac{(1-1/P)}{K^2}
    \end{split}
\end{equation}
In the Equation-\ref{eq:speedup} if we put $P=1$ then it becomes standard convolutional filter. 

By reducing the size of the filter on some channels from says $3\times 3$ to $1\times 1$, we are reducing the spatial extent of a filter. However, by retaining the size to be $3\times 3$ on some channel, we ensure that the filter does cover the spatial correlation on some channels and need not to have the same spatial correlation on all channels. We observe in the experimental section that by doing so, one can obtain similar accuracies as a homogeneous filter. On the other hand, if we avoid and retain a $1\times 1$ filter size on all channels, then we would not have the necessary spatial correlation information covered, and the accuracy would suffer.

\subsection{Comparision with DepthWise followed by PointWise Convolution}
In extreme case when $P=M$ in HetConv, HetConv can be compared with DWC+PWC (DepthWise followed by PointWise Convolution). MobileNet \cite{howard2017mobilenets} use this type of convolution.  While MobileNet takes more FLOPs than our extreme case with the more delay since MobileNet \cite{howard2017mobilenets} has latency one. 

The total FLOPs for DWC+PWC (MobileNet) for layer L can be computed as:
\begin{equation}
    FL_{MobNet}=D_o \times D_o \times M  \times K \times K + M  \times N \times D_o \times D_o
\end{equation}
Therefore the total reduction in computation as compared to the standard convolution:
\begin{equation}\label{eq:speedupmob}
    \begin{split}
    R_{MobNet}=&\frac{FL_{MobNet}}{FL_S}\\
    =&\frac{1}{N}+\frac{1}{K^2}
    \end{split}
\end{equation}

It is clear from the Equation-\ref{eq:speedup} that we can change the part $P$ value to trade off between the accuracy and FLOPs. If we decrease the $P$ value, the resulting convolution will be closer to the standard convolution. To show the effectiveness of the proposed HetConv filter, we have shown results in the experimental section where HetConv achieves significantly better accuracy with similar FLOPs.

In the extreme case when $P=M$, from Eq.-\ref{eq:speedup} and \ref{eq:speedupmob} (for MobileNet $N=M$),we can conclude:
\begin{equation}\label{eq:mobilecomp}
    \frac{1}{M}+\frac{(1-1/M)}{K^2} < \frac{1}{M}+\frac{1}{K^2}
\end{equation}

Reduction = Total reduction in the computation as compared to standard convolution

\begin{equation}
    Speedup = \frac{1}{Reduction}
\end{equation}


Therefore from Eq.-\ref{eq:mobilecomp}, it is clear that MobileNet takes more computation than our approach. In our HetConv, we have latency zero while MobileNet has latency one. In this extreme case, we have significantly better accuracy than MobileNet (refer to the experiment section).

\subsection{Comparision with GroupWise followed by PointWise Convolution}
In the case of groupwise convolution followed by pointwise convolution (GWC+PWC) with the group size $G$, the total FLOPs for GWC+PWC for layer L can be computed as:
\begin{equation}
    FL_{G}=(D_o \times D_o \times M \times N \times K \times K)/G + M  \times N \times D_o \times D_o
\end{equation}
Therefore the total reduction in the computation as compare to the standard convolution:
\begin{equation}\label{eq:speedupG}
    \begin{split}
    R_{Group}=&\frac{FL_{G}}{FL_S}\\
    =&\frac{1}{G}+\frac{1}{K^2}
    \end{split}
\end{equation}

Similarly when $P=G$, from Eq.-\ref{eq:speedup} and \ref{eq:speedupG}  we have:
\begin{equation}\label{eq:speedupGpf}
    \frac{1}{P}+\frac{(1-1/P)}{K^2} < \frac{1}{P}+\frac{1}{K^2}
\end{equation}
Therefore from Eq.-\ref{eq:speedupGpf}, it is clear that GWC+PWC takes more computation than our approach. In our HetConv, we have latency zero while GWC+PWC has latency one.

\subsection{Running Latency}

\begin{figure}[t]
    \centering
    \includegraphics[scale=0.37]{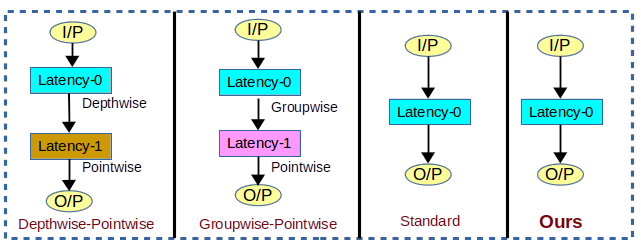}
    \caption{The figure shows the comparison with the different types of convolution in terms of latency.}
    \label{fig:latency}
\end{figure}

Most of the previous approaches \cite{szegedy2017inception,szegedy2015googlenet,zhangshufflenet,howard2017mobilenets} designed efficient convolution to reduce the FLOPs, but they increase the latency\footnote{One parallel step is converted to multiple sequential step hence reduction in parallelizability. Later stage of layers waits for the execution to be finished on the previous stage because all computations have to be done sequentially across layers} in the architecture. The latency in the different types of convolutions is shown in the Figure-\ref{fig:latency}. In the Inception module \cite{szegedy2015googlenet,szegedy2017inception}, one layer is broken down into two or more sequential layers. Therefore, the latency in the architecture is greater than zero. In the Xception \cite{chollet2017xception} first GWC is applied, and on the output of GWC, PWC is applied. PWC waits for the completion of the GWC. Hence this approach reduces the FLOPs but increases latency in the system.
Similarly in the MobileNet \cite{howard2017mobilenets} first DWC and then PWC is applied therefore it has latency one. This latency includes a delay in parallel devices like GPU. In our proposed approach any layers are not replaced by sequential layers hence has the latency zero. We directly design the filter such that without increasing any latency we can reduce the FLOPs. Our proposed approach is very competitive in terms of FLOPs as compared to previous efficient convolutions while maintaining the latency zero.

\subsection{Speedup over standard convolution for different values of $P$}
\begin{figure}
    \centering
    \includegraphics[height=5.2cm,width=8.5cm]{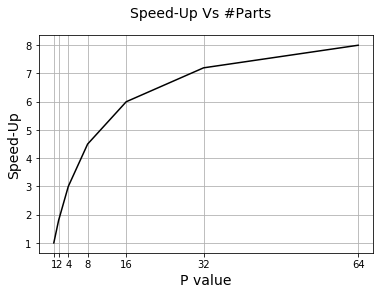}
    \caption{Speedup over standard convolution for different values of $P$ for a HetConv Filter with $3\times 3$ and $1\times 1$ kernels.}
    \label{fig:my_label}
\end{figure}
As shown in Figure-\ref{fig:my_label}, Speedup increases with the $P$ value. We can use $P$ value to trade off between the accuracy and FLOPs. If we decrease the $P$ value, the resulting convolution will be closer to the standard convolution. To show the effectiveness of the proposed HetConv filter, we have shown results in the experimental section where HetConv achieved significantly better accuracy with respect to the other types of convolution with similar FLOPs.

\section{Experiments and Results}
To show the effectiveness of the proposed HetConv filter we perform  extensive experiments with the current state-of-art architectures. We replaced their standard convolutional filters from these architecture with the proposed one. We performed three large scale experiment on the ImageNet \cite{imagenet2015} with the ResNet-34, ResNet-50 \cite{resnet} and VGG-16 \cite{vgg2014very} architectures. We have shown three small scale experiment on the CIFAR-10 \cite{cifar10} for the VGG-16, ResNet-56, and MobileNet \cite{howard2017mobilenets} architectures. 
We set the value of reduction ratio to 8 for Squeeze-and-Excitation (SE) \cite{hu2017squeeze} in all our experiments.

\subsection{Notations}
XXX\_P$\alpha$: XXX is the architecture, and part value is $P=\alpha$; XXX\_P$\alpha$\_SE: SE for Squeeze-and-Excitation with reduction-ratio = 8; XXX\_GWC$\beta$\_PWC: GWC$\beta$\_PWC is the groupwise convolution with group size $\beta$ followed by pointwise convolution; XXX\_DWC\_PWC: DWC\_PWC is depthwise convolution followed by pointwise convolution; XXX\_PC: PC is part value $P$ = number of input channels (input depth).  

\subsection{VGG-16 on CIFAR-10}

\begin{table*}[!hbt]
    \centering
    \scalebox{0.9}{
        \addtolength{\tabcolsep}{7pt}
        \begin{tabular}{|l| c| c| c| c|c|c|}
            \hline
            Model & Acc\% & FLOPs & FLOPs Reduced (\%) &Parameters & Parameters Reduced (\%)\\ 
            \hline \hline
            
            VGG-16\_P1 & 94.06 & 313.74M & -- & 15.00M&-- \\
            VGG-16\_P1\_SE & 94.13 & 314.19M & -- & 15.22M &--\\
            \hline \hline
            VGG-16\_P2 & 93.89 & 175.23M & 44.15 & 8.45M & 43.68\\
            VGG-16\_P2\_SE & 94.11 & 175.67M & 44.00 & 8.68M & 42.99\\
            \hline \hline
            VGG-16\_P4 & 93.93 & 105.98M & 66.22 & 5.17M & 65.45\\ 
            VGG-16\_P4\_SE & 94.29 & 106.42M & 66.08 & 5.41M & 64.48\\
            VGG-16\_GWC4\_PWC & 92.76 & 107.67M & 65.68 & 5.42M & --\\
            \hline \hline
            VGG-16\_P8 & 93.92 & 71.35M & 77.26 & 3.54M & 76.40\\
            VGG-16\_P8\_SE & 93.97 & 71.79M & 77.12 & 3.77M & 75.22\\
            \hline \hline
            VGG-16\_P16 & 93.96 & 54.04M & 82.78 & 2.72M & 81.86\\
            VGG-16\_P16\_SE & 93.63 & 54.48M & 82.64 & 2.95M & 80.59\\
            \hline \hline
            VGG-16\_P32 & 93.73 & 45.38M & 85.54 & 2.31M & 84.58 \\
            VGG-16\_P32\_SE & 93.41 & 45.82M & 85.39 & 2.54M & 83.28\\
            \hline \hline
            VGG-16\_P64 & 93.42 & 41.05M & 86.92 & 2.11M & 85.95\\
            VGG-16\_P64\_SE & 93.33 & 41.49M & 86.77 & 2.34M & 84.63\\
            \hline \hline
            VGG-16\_DWC\_PWC & 91.27 & 38.53M & 87.72 & 1.97M & --\\
            VGG-16\_PC & 92.53 & 38.18M & 87.83& 1.93M & --\\
            VGG-16\_PC\_SE & 93.08 & 38.62M & 87.69 & 2.15M & --\\ \hline
            
        \end{tabular} 
    }
    \vspace{2pt}
    \caption{The table shows the detail results for VGG-16 on CIFAR-10 in different setups.}
    \label{tab:vggall}
\end{table*}

In this experiment, we use the VGG-16 architecture \cite{vgg2014very}. In the CIFAR-10 dataset, each image size is of $32\times32$ size on RGB scale. In VGG-16 architecture, there are 13 convolutional layers which use standard CONV2D convolution, and after each layer, we add batch normalization. We are using the same setting as described in \cite{li2016pruning}. The values of hyper-parameters are: weight decay = 5e-4, batch size = 128, initial learning rate = 0.1 which is decade by 0.1 after every 50 epochs.

Except for the initial convolution layer (i.e., CONV\_1), All remaining 12 standard convolutional layers are replaced by our HetConv layers (same $P$ value for all 12 layers) while keeping the number of filers same as earlier. As shown in Table-\ref{tab:vggall}, the value of $P$ increase, the values of FLOPs (computation) decreases without any significant drop in accuracy. We also experimented for HetConv with SE technique and found that SE increases the accuracy initially but later due to over-fitting, it starts degrading the model performance (accuracy) as shown in Table-\ref{tab:vggall}. 

\subsubsection{Comparison with groupwise followed by pointwise convolution}
We experimented with groupwise followed by pointwise convolution, where all standard convolutional layers (except the initial convolution layer, i.e., CONV\_1) are replaced by two layers (groupwise convolutional layer with group size 4 and pointwise convolutional layer). Now the model has latency one. As shown in Table-\ref{tab:vggall}, VGG-16\_GWC4\_PWC has 92.76\% accuracy whereas our model VGG-16\_P4 has significantly higher 93.93\% accuracy with lesser FLOPs. 
\subsubsection{Comparison with depthwise followed by pointwise convolution}
We experimented with depthwise followed by pointwise convolution, where all standard convolutional layers (except the initial convolution layer, i.e., CONV\_1) are replaced by two layers (depthwise convolutional layer and pointwise convolutional layer). Now the model has latency one. As shown in Table-\ref{tab:vggall}, VGG-16\_DWC\_PWC has only 91.27\% accuracy on CIFAR-10 whereas our model VGG-16\_P64 has significantly higher 93.42\% accuracy with comparable FLOPs. 

We also experimented with different $P$ values for different layers. Except for the initial convolution layer (i.e., CONV\_1), all remaining 12 standard convolutional layers are replaced by our HetConv layers with $P = \text{number of input channels}$. As shown in Table-\ref{tab:vggall}, our model VGG-16\_PC and VGG-16\_PC\_SE still performing better than VGG-16\_DWC\_PWC which shows the superior performance of our HetConv over DWC+PWC. 

\subsubsection{Comparison with FLOPs compression methods}

\begin{table}[!ht]
    \centering
    \scalebox{0.92}{
    \addtolength{\tabcolsep}{3pt}
    \begin{tabular}{|l| c| c| c|} 
    \hline
    Method &  Error\%  & FLOPs Reduced(\%) \\
    \hline\hline
    Li-pruned \cite{weightSumICLR-17} &  6.60 & 34.20\\
    SBP \cite{neklyudov2017structured} &  7.50 &  56.52\\
    SBPa \cite{neklyudov2017structured} &  9.00 & 68.35\\
    AFP-E \cite{ding2018auto} &  7.06 &  79.69\\
    AFP-F \cite{ding2018auto} & 7.13 & 81.39\\
    \hline\hline
    \textbf{VGG-16\_P32 (Ours)} &  \textbf{6.27} &   \textbf{85.54} \\
    \textbf{VGG-16\_P64 (Ours)} & \textbf{6.58} &  \textbf{86.92} \\
    \hline
    \end{tabular}
    }
    \vspace{2pt}
    \caption{The table shows the comparison of our models with state-of-art model compression methods for VGG-16 architecture on the CIFAR-10 dataset.}
    \label{tab:vgglike}
\end{table}

As shown in Table-\ref{tab:vgglike}, our models VGG-16\_P32, and VGG-16\_P64 have significantly better accuracy as compare to state-of-art model compression methods. We reduced $\sim$ 85\% FLOPs with no loss in accuracy whereas FLOPs compression methods suffer a significant loss in accuracy (more than 1\%) as shown in Table-\ref{tab:vgglike}.
\subsection{ResNet-56 on CIFAR-10}

\begin{table}[b]
    \centering
    \scalebox{0.92}{
    \addtolength{\tabcolsep}{5pt}
    \begin{tabular}{|l| c| c| c|} 
    \hline
    Method &  Error\%  & FLOPs Reduced (\%) \\
    \hline\hline
    Li-B \cite{weightSumICLR-17} &  6.94 &  27.6\\
    NISP \cite{yu2017nisp} &  6.99 &  43.6\\
    CP \cite{channelPruning17} &  8.20 &  50.0\\
    SFP \cite{he2018soft} &  6.65 & 52.6\\
    AFP-G \cite{ding2018auto} &  7.06 &  60.9\\
    \hline\hline
    {ResNet-56\_P1} &  6.41 &   -- \\
    \textbf{ResNet-56\_P2} &  \textbf{6.40} &   \textbf{44.30} \\
    \textbf{ResNet-56\_P4} &  \textbf{6.71} &   \textbf{66.45} \\
    \hline\hline
    {ResNet-56\_P1\_SE} &  7.16 &   -- \\
    {ResNet-56\_P2\_SE} &  {6.75} &   {44.27} \\
    {ResNet-56\_P4\_SE} &  {7.79} &   {66.42} \\
    \hline
    \end{tabular}
    }
    \vspace{2pt}
    \caption{The table shows the detail results and comparison with state-of-art model compression methods for ResNet-56 on CIFAR-10 in different setups.}
    \label{tab:resnet56}
\end{table}

We experimented with ResNet-56 architecture \cite{resnet} on the CIFAR-10 dataset. ResNet-56 consists of three stages of the convolutional layer of size 16-32-64 where each convolution layer in each stage contains the same 2.36M FLOPs, and the total FLOP is 126.01M. We trained the model using the same parameters proposed by \cite{resnet}. Except for the initial convolution layer, all remaining standard convolutional layers are replaced by our HetConv layers while keeping the number of filers same as earlier. 

As shown in Table-\ref{tab:resnet56}, our models ResNet-56\_P2, and ResNet-56\_P4 have significantly better accuracy as compare to state-of-art model compression methods with higher FLOPs reduction. We also experimented for HetConv with SE technique and found that SE performance is not as expected due to over-fitting. 

\subsection{MobileNet on CIFAR-10}

\begin{table}[t]
    \centering
    \scalebox{1.0}{
    \addtolength{\tabcolsep}{5pt}
    \begin{tabular}{|l| c| c| c|} 
    \hline
    Method &  Accuracy (\%)  & FLOPs \\
    \hline\hline
    MobileNet \cite{howard2017mobilenets} &  91.17 &  46.36M\\
    
    \textbf{MobileNet\_P32} &  \textbf{92.06} &   {55.94M} \\
    \textbf{MobileNet\_P32\_SE} &  \textbf{92.17} &  {56.91M} \\
    \hline
    \end{tabular}
    }
    \vspace{2pt}
    \caption{The table shows the results for MobileNet \cite{howard2017mobilenets} on CIFAR-10 in different setups.}
    \vspace{-15pt}
    \label{tab:mobile}
\end{table}

We experimented with MobileNet architecture on the CIFAR-10 dataset. Except for the initial convolution layer, all remaining convolutional layers are replaced by our HetConv layers while keeping the number of filers same as earlier. In our model, two convolutional layers (depthwise convolutional layer and pointwise convolutional layer) is replaced by one HetConv convolutional layer which reduces the latency from one to zero. 

As shown in Table-\ref{tab:mobile}, our models MobileNet\_P32, and MobileNet\_P32\_SE have the significantly better accuracy (close to 1\%) as compare to MobileNet model with almost similar FLOPs on MobileNet architecture which again shows the superior performance of our proposed HetConv convolution over depthwise+pointwise convolution.

\subsection{VGG-16 on ImageNet}

\begin{table}[!h]
    \centering
    \scalebox{0.7}{
    \addtolength{\tabcolsep}{5pt}
    \begin{tabular}{|l| c| c| c| c|}
    \hline
    Method & Acc\%(Top-1) & Acc\%(Top-5) & FLOPs Reduced \% \\ [0.8ex] 
    \hline\hline
    RNP (3X)\cite{lin2017runtime} &  -- &  87.57 & 66.67\\
    ThiNet-70 \cite{thiNet17} &  69.8 &  89.53 & 69.04\\
    CP 2X \cite{channelPruning17} &  -- &  89.90 & 50.00\\ 
    \hline\hline
    VGG-16\_P1 & 71.3 & 90.2 & --\\
    \textbf{VGG-16\_P4} & \textbf{71.2} & \textbf{90.2} & \textbf{65.8}\\
    \hline
    \end{tabular} }
    \vspace{2pt}
    \caption{Table shows the results for the VGG-16 on ImageNet \cite{imagenet2015}. Our model has no loss in accuracy as compare to state-of-art \cite{channelPruning17,thiNet17} pruning approaches while significantly higher FLOPs reduction.}
    \label{tab:vgg16Imagenet}
\end{table}

We experimented with VGG-16 \cite{vgg2014very} architecture on the large-scale ImageNet \cite{imagenet2015} dataset. Except for the initial convolution layer, all remaining convolutional layers are replaced by our HetConv layers while keeping the number of filers same as earlier. Our model VGG-16\_P4 shows the state-of-art result over the recent approach proposed for flop compression. Channel-Pruning (CP) \cite{channelPruning17} has the $50.0\%$ FLOP compression while we have $65.8\%$ FLOP compression with no loss in accuracy. Please refer to Table-\ref{tab:vgg16Imagenet} for the more detail results.

\subsection{ResNet-34 on ImageNet}

\begin{table}[!ht]
    \centering
    \scalebox{.9}{
    \addtolength{\tabcolsep}{-4pt}
    \begin{tabular}{|l| c| c| c|} 
    \hline
    Method &  Error (top-1)\% & FLOPs  & FLOPs Reduced(\%) \\
    \hline\hline
    Li-B \cite{weightSumICLR-17} & 27.83 & 2.7G & 24.2\\
    NISP \cite{yu2017nisp} & 27.69 & -- & 43.76\\
    \hline\hline
    {ResNet-34\_P1} & 26.80 & 3.6G &-- \\
    \textbf{ResNet-34\_P4} &  \textbf{27.00} & 1.3G & \textbf{64.48}
    \\
    \textbf{ResNet-34\_P4\_SE} &  \textbf{26.50} & 1.3G & \textbf{64.48}\\
    \hline
    \end{tabular}
    }
    \vspace{2pt}
    \caption{Table shows the results for ResNet-34 on ImageNet \cite{imagenet2015}. Our model has no loss in accuracy as compare to state-of-art \cite{weightSumICLR-17,yu2017nisp} pruning approaches while significantly higher FLOPs reduction in different setups.}
    \label{tab:resnet34}
\end{table}

We experimented with ResNet-34 \cite{resnet} architecture on the large-scale ImageNet \cite{imagenet2015} dataset. Except for the initial convolution layer, all remaining convolutional layers are replaced by our HetConv layers. Our model ResNet-34\_P4 shows the state-of-art result over the previously proposed methods. NISP \cite{yu2017nisp} has the $43.76\%$ FLOP compression while we have $64.48\%$ FLOP compression with significantly better accuracy. For more details, please refer to Table-\ref{tab:resnet34}.

\subsection{ResNet-50 on ImageNet}

\begin{table}[!ht]
    \centering
    \scalebox{0.9}{
    \addtolength{\tabcolsep}{-3pt}
    \begin{tabular}{|l| c| c| c|} 
    \hline
    Method &  Error (top-1)\% & FLOPs  & FLOPs Reduced(\%) \\
    \hline\hline
    ThiNet-70 \cite{thiNet17} & 27.90 & -- & 36.8\\
    NISP \cite{yu2017nisp} & 27.33 & -- & 27.31\\
    
    \hline\hline
    {ResNet-50\_P1} & 23.86 & 4.09G &-- \\
    \textbf{ResNet-50\_P4} &  \textbf{23.84} & 2.85G & \textbf{30.32} \\
    \hline
    \end{tabular}
    }
     \vspace{2pt}
    \caption{Table shows the results for ResNet-50 on ImageNet \cite{imagenet2015}. Our model has no loss in accuracy as compare to state-of-art \cite{thiNet17,yu2017nisp} flop pruning approaches.}
    \label{tab:resnet50}
\end{table}

ResNet-50 \cite{resnet} is a deep convolutional neural network having 50 layers with the skip/residual connection. In this architecture, we replace standard convolutions with our proposed HetConv convolution. The values of hyper-parameters are: weight decay = 1e-4, batch size = 256, initial learning rate = 0.1 which is decade by 0.1 after every 30 epochs and model is trained in 90 epochs.

It is clear from Table-\ref{tab:resnet50} that our model ResNet-50\_P4 has no loss in accuracy while flop pruning approaches \cite{thiNet17,yu2017nisp} suffers from the heavy accuracy drop in top-1 accuracy. Our model is trained from scratch, but pruning approaches \cite{thiNet17,yu2017nisp} requires a pre-trained model and involve iterative pruning and fine-tuning which is a very time-consuming process.   


\section{Conclusion}
In this work, we proposed a new type of convolution using heterogeneous kernels. We have compared our proposed convolution with the popular convolutions (depthwise convolution, groupwise convolution, pointwise convolution, and standard convolution) on various existing architectures (VGG-16, ResNet and MobileNet). Experimental results show that our HetConv convolution is more efficient (lesser FLOPs with better accuracy) as compared to existing convolutions. Since our proposed convolution does not increase the layer (replacing a layer with a number of layers, for example, MobileNet) to get FLOPs reduction, hence has latency zero. We also compared HetConv convolution based model with the FLOPs compression methods and shown that it produces far better results as compared to compression methods. In the future, using this type of convolution, we can design more efficient architectures.

\newpage
{\small
\bibliographystyle{ieee}
\bibliography{egbib}
}

\end{document}